\documentclass{article}
\usepackage{spconf,amsmath,graphicx,hyperref}

\title{When Forgetting Looks Like Improvement: Metric Masking in Streaming Diarizer Adaptation and the Price of Rehearsal}
\name{Mo Yu, Yang Liu$^{*}$, Jing Qian$^{*}$%
\thanks{$^{*}$Co-corresponding authors: Yang Liu (yangliu25@tongji.edu.cn) 
and Jing Qian (jqian1590@tongji.edu.cn).}}

\address{Tongji University\\
Shanghai, China\\
moy97021@gmail.com, yangliu25@tongji.edu.cn, jqian1590@tongji.edu.cn}
\begin{document}
%\ninept
%
\maketitle
\begin{abstract}
Small-data adaptation can improve speech detection while degrading speaker attribution. We study this discrepancy in a released streaming diarizer adapted on 7.5 h of two-party conversation and evaluated across six corpora. Adaptation substantially improves in-domain diarization performance and transfers to an independent corpus. However, this improvement is not consistent across evaluation scenarios as the additional confusion is mainly associated with impaired temporal identity consistency rather than speaker-count errors. A local-remapping diagnostic reveals different patterns of identity degradation across corpora, indicating that adaptation may alter how streaming models maintain speaker assignments over time. Rehearsal reduces the observed degradation but reduces the cross-domain transfer performance. These results highlight the need to jointly evaluate detection accuracy, identity consistency, and retention behavior when adapting streaming diarization systems.
\end{abstract}
\section{Introduction}
\label{sec:intro}

Streaming speaker diarization must detect speaker activity and maintain consistent identities as audio arrives~\cite{park2022review}. When combining with a Streaming Sortformer it addresses online identity tracking through a speaker cache~\cite{medennikov2025streaming}. Fine-tuning such a pretrained model for a target domain raises an important question of \textbf{what improves, and what retains}? While Aggregate diarization error rate (DER) is commonly used to indicate performance, the questions remains unclear as gains sometimes hide other attributes such as increased speaker confusion.

Prior work has investigated continual self-supervised domain adaptation~\cite{coria2023continual}. Our focus is on how these adaptation changes detection, speaker assignment, and temporal consistency across domains~\cite{lanzendoerfer2025benchmarking}. We conduct a diagnostic study of Streaming Sortformer v2, comparing the released checkpoint with target-only adaptation and mixed-data rehearsal variants across six corpora. We first decompose DER, then equalize framewise active-output counts to compare assignment under identical scored detection errors. Short-window speaker remapping relaxes recording-wide correspondence without changing predictions. History interventions further probe sensitivity to streaming state, while the rehearsal comparison examines recovery and sacrifices.

Our contribution is threefold. First, we demonstrate a gap between score-level retention and speaker-assignment retention; second, we uncover history-sensitive consistency costs beyond short-window confusion, revealing a temporal dimension of adaptation that aggregate scores obscure; and third, the tested rehearsal recipe redistributes retention and transfer, as aggregate recovery comes with lost transfer gains, and near-baseline DER can still conceal assignment degradation. 

\section{Experimental protocol}
\label{sec:protocol}

\subsection{Models, adaptation, and evaluation data}

The released \texttt{diar\_streaming\_\allowbreak sortformer\_4spk-v2} model~\cite{nvidia2025sortformerv2} (\textbf{Stock S}) has four speaker slots ordered by first appearance~\cite{park2025sortformer} and produces activity scores at 12.5 Hz. \textbf{Expert (E)} fine-tunes all 117 M parameters for 24 epochs using AdamW~\cite{loshchilov2019adamw}, learning rate $10^{-5}$, batches of four 60-s crops, oversampling factor two, gradient clipping at 5.0, and an auxiliary Dice-loss weight of 0.2 in NeMo~\cite{kuchaiev2019nemo}. We use the final-epoch checkpoint. \textbf{Generalist (G)} adds rehearsal~\cite{chaudhry2019tiny,rolnick2019replay}~\ref{sec:rehearsal}. All models use mono channel 0 and the same released inference implementation at the 1.04-s latency configuration. 

The model is fine-tuned on MMCSG dataset~\cite{zmolikova2024mmcsg}, its 's official evaluation partition is re-split into 153 training sessions (7.5 h) and 36 speaker-disjoint holdout sessions (1.8 h). The current training and holdout sets are disjoint, but the holdout participated in earlier recipe development. Eight sessions were in the earlier validation split and 28 in its training split. Aside from in-domain evaluation, we also evaluate PriMock57~\cite{korfiatis2022primock57} and EasyCom~\cite{donley2021easycom}. Development corpora including MSDWild~\cite{liu2022msdwild} few.val, AliMeeting-Test~\cite{yu2022m2met}'s two-speaker subset, and AMI~\cite{carletta2006ami} single-distant-microphone test were used for informed rehearsal development. 

\begin{table*}[t]
\centering
\caption{Adaptation and retention under matched count FA. S/E/G denote Stock/Expert/Generalist.}
\label{tab:main}

\small
\setlength{\tabcolsep}{3pt}

\begin{tabular}{llcrrrrrrr}
\hline
Corpus & Role & Model & Count FA &
gMiss & ov-Miss & DER miss &
DER FA & Confusion & DER \\
\hline

MMCSG (36) & ID & S & 2.612 &
12.36 & 33.75 & 12.37 & 2.61 & 0.59 & 15.57 \\

& & E & 2.618 &
4.93 & 22.13 & 4.95 & 2.62 & 0.25 & 7.82 \\

& & G & 2.620 &
4.88 & 24.02 & 4.90 & 2.62 & 0.31 & 7.83 \\

\hline

PriMock57 (57) & I & S & 2.582 &
15.86 & 41.77 & 15.87 & 2.59 & 0.68 & 19.15 \\

& & E & 2.593 &
5.43 & 26.34 & 5.46 & 2.62 & 0.34 & 8.42 \\

& & G & 2.613 &
15.36 & 37.83 & 15.37 & 2.63 & 1.40 & 19.40 \\

\hline

EasyCom (223) & I & S & 2.577 &
31.28 & 50.88 & 32.80 & 2.46 & 7.34 & 42.60 \\

& & E & 2.562 &
28.31 & 49.26 & 30.09 & 2.44 & 10.86 & 43.39 \\

& & G & 2.526 &
27.16 & 49.76 & 28.89 & 2.42 & 8.56 & 39.87 \\

\hline

MSDWild (274) & D & S & 2.608 &
17.18 & 40.91 & 17.20 & 2.64 & 1.88 & 21.72 \\

& & E & 2.721 &
22.71 & 45.35 & 22.72 & 2.74 & 1.25 & 26.71 \\

& & G & 2.593 &
15.10 & 34.28 & 15.12 & 2.63 & 2.10 & 19.85 \\

\hline

AliMeeting-2 (8) & D & S & 2.631 &
3.07 & 9.93 & 3.11 & 2.66 & 0.12 & 5.90 \\

& & E & 2.621 &
6.49 & 33.99 & 6.51 & 2.63 & 0.11 & 9.25 \\

& & G & 2.617 &
2.77 & 8.62 & 2.81 & 2.65 & 0.13 & 5.59 \\

\hline

AMI (16) & D & S & 2.627 &
23.70 & 35.91 & 24.90 & 2.59 & 7.75 & 35.23 \\

& & E & 2.629 &
20.15 & 40.79 & 22.87 & 2.49 & 21.46 & 46.82 \\

& & G & 2.654 &
15.89 & 34.44 & 17.98 & 2.54 & 8.58 & 29.10 \\

\hline

\end{tabular}
\end{table*}

% All printed material, including text, illustrations, and charts, must be kept
% within a print area of 7 inches (178 mm) wide by 9 inches (229 mm) high. Do
% not write or print anything outside the print area. The top margin must be 1
% inch (25 mm), except for the title page, and the left margin must be 0.75 inch
% (19 mm).  All {\it text} must be in a two-column format. Columns are to be 3.39
% inches (86 mm) wide, with a 0.24 inch (6 mm) space between them. Text must be
% fully justified.

\subsection{Metric scope and operating points}
\label{sec:metrics}

For frame $t$, let $r_t$ and $h_t$ count reference speakers and active output slots. On scoring frames $\mathcal{T}$, count-based global Miss and false alarm are
\begin{equation}
\begin{aligned}
 M(\mathcal{T})&=\frac{100\sum_{t\in\mathcal{T}}(r_t-h_t)_+}{\sum_{t\in\mathcal{T}}r_t},\\
 F(\mathcal{T})&=\frac{100\sum_{t\in\mathcal{T}}(h_t-r_t)_+}{\sum_{t\in\mathcal{T}}r_t}.
\end{aligned}
\label{eq:count}
\end{equation}
Within-overlap Miss (ov-Miss) restricts $\mathcal{T}$ to $r_t\ge2$. Both miss measures ignore speaker identity. For AMI and EasyCom, count metrics and threshold calibration additionally restrict $\mathcal{T}$ to $r_t\le2$. Standard DER uses the full continuous reference~\cite{ryant2020third}. Count Miss and DER miss are not interchangeable. A separate diagnostic recomputes DER within the count mask, with its own optimal mapping.

For each corpus, we select Stock's threshold nearest a nominal count-FA target of 2.62\%, then select each adapted model's threshold nearest Stock's achieved FA on a 0.005 grid. Table~\ref{tab:main} reports actual count FA. All thresholds use labels from the scored corpus, including the two recipe-independent corpora. These are controlled diagnostic operating points. Sensitivity analyses use 2.0\% and 3.5\% targets and grouped two-fold calibration on AMI and EasyCom.

\subsection{Uncertainty and replication}
\label{sec:stats}
%Corpus-level metric differences use unrounded pooled error totals with paired group resampling. 
We use nine speaker-connected groups for MMCSG, ten parent sessions for EasyCom, and recording units for PriMock57 (57), MSDWild (274), and AliMeeting-2 (8), with 20,000 resamples. For AMI, we group recordings into four meeting series and enumerate all 256 bootstrap sequences. Intervals are conditional on the checkpoints, thresholds, and grouping; they do not resolve unverified participant dependencies or training uncertainty and are not adjusted for multiple comparisons. Three Expert seeds are evaluated on AMI, EasyCom, and PriMock57. 

\begin{figure}[t]
\centering
\includegraphics[width=0.98\columnwidth]{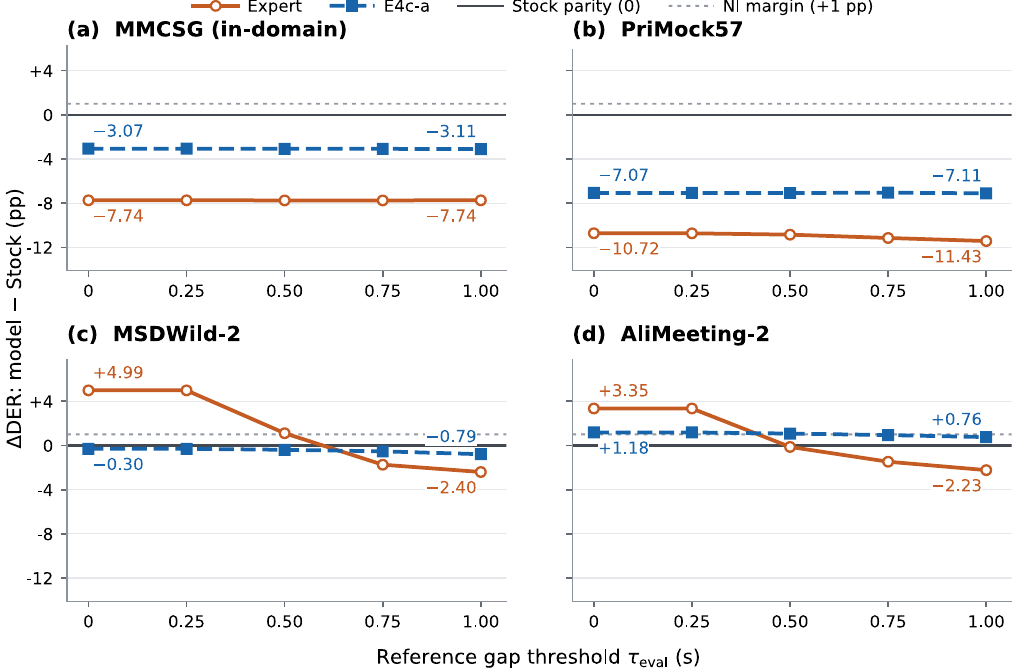}
\caption{
Changing of convention-aligned labels lower the DER on MSDWild-2 and Alimeeting-2 datasets but remain unchanged for in-domain and PriMock57.
}
\label{fig:tau-analysis}
\end{figure}

\section{Adaptation improves detection while altering identity retention}
\label{sec:masking}

\subsection{Adaptation with heterogeneous retention}

\begin{figure}[t]
\centering
\includegraphics[width=0.75\columnwidth]{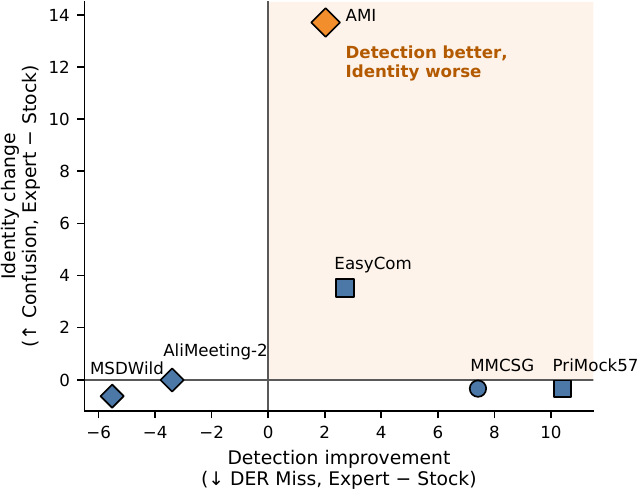}
\vspace{-1mm}
\includegraphics[width=0.75\columnwidth]{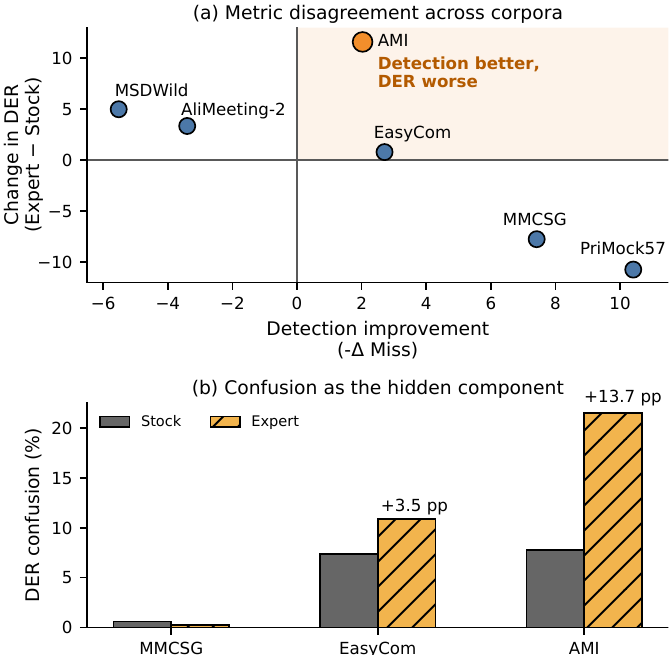}
\caption{
Adaptation changes diarization capabilities beyond aggregate DER.
Detection improvements can coexist with increased speaker confusion,
which is hidden when only aggregate DER is considered.
}
\label{fig:adaptation_tradeoff}
\end{figure}

%Adaptation does not improve diarization along a single dimension: aggregate DER combines gains in speech activity detection with changes in speaker identity assignment. 
Expert reduces MMCSG ov-Miss from 33.75\% to 22.13\% and DER from 15.57\% to 7.82\% (Table~\ref{tab:main}). The paired DER change is $-7.74$ pp [$-9.73,-5.35$]. On recipe-independent PriMock57, ov-Miss improves by 15.43 pp and DER by 10.72 pp [$-11.94,-9.65$]. These gains show substantial adaptation headroom. Retention is domain-dependent: DER increases by 4.99 pp on MSDWild and 3.35 pp on AliMeeting-2, predominantly through missed detection, while AMI and EasyCom exhibit a different error composition.

\subsection{Detection gains can conceal identity degradation}

On AMI, full-scope DER miss falls from 24.90\% to 22.87\% and DER FA from 2.59\% to 2.49\%, yet confusion increases from 7.75\% to 21.46\%. %Consequently, DER rises by 11.59 pp. The 13.71-pp confusion increase outweighs the detection gains. Count gMiss also falls by 3.55 pp, but its series-grouped interval [$-10.00,+2.62$] includes zero. The evidence for increased identity error is  stronger than the evidence for a population-level detection improvement.

EasyCom provides independent corroboration of the identity deficit. Expert lowers gMiss by 2.97 pp [$-5.62,-0.04$] while increasing confusion by 3.52 pp [$+2.62,+4.47$]. Overall DER changes by $+0.79$ pp [$-1.53,+3.39$]. This distinction shows that detection gains partly compensate for an identity loss that the decomposed evaluation detects. Scoring DER within the same $r_t\le2$ mask preserves the point-estimate reversal on both corpora (AMI $+11.03$ pp; EasyCom $+0.67$ pp). Scoring scopes alone do not explain the phenomenon.

AliMeeting-2 illustrates a complementary effect as its 24.06-pp ov-Miss increase corresponds to only 3.35 pp of DER degradation. Decomposing the 3.42-pp increase in global count Miss attributes 2.56 pp (74.9\%) to overlap and 0.86 pp to non-overlap. Overlap-restricted scoring exposes a concentrated weakness that whole-recording averaging reduces in magnitude.

\subsection{Replication and calibration sensitivity}
The replications support the direction of the principal domain contrasts without treating seeds as additional evaluation samples. Across three Expert seeds, dataset exhibits decreased gMiss while increased confusion and DER. At nominal count-FA targets of 2.0\%, 2.62\%, and 3.5\%, Expert has lower gMiss but higher DER on both AMI and EasyCom. The DER increases range from 11.31 to 11.59 pp and from 0.27 to 1.52 pp, respectively. Grouped two-fold calibration preserves this direction in pooled held-out results for all three distinct balanced AMI partitions and five EasyCom partitions.% Held-out FA is not exactly matched, and these repeated partitions use the same data. 
The checks establish persistence over the examined operating points, not threshold-invariant effect sizes or significance in every fold.

\section{Diagnosing the identity deficit}
\label{sec:identity}

\subsection{Local assignment and recording-wide consistency}

We retain predictions and compare recording-wide speaker mapping with independent optimal mappings over non-overlapping windows of length $W$. Let $C$ be pooled confusion with one optimal mapping per recording and $L_W$ its locally mapped counterpart, using identical time support and reference speaker-time. Define:
$ C=L_W+G_W, G_W=C-L_W\ge0.$

\begin{figure}[t]
\centering
\includegraphics[width=0.75\columnwidth]{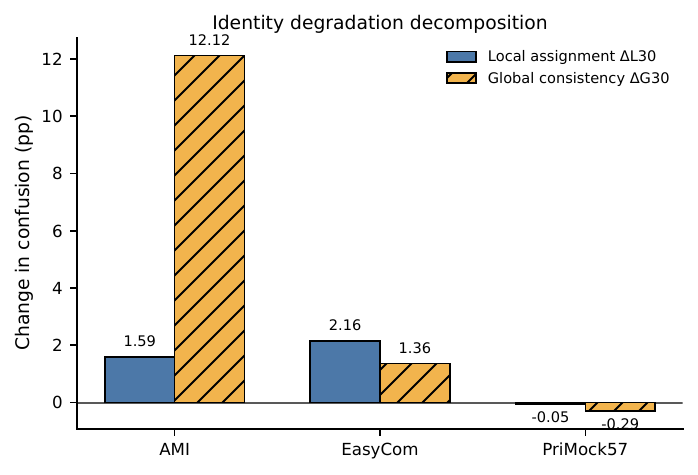}
\caption{
Decomposition of increased confusion into local assignment error and recording-wide consistency penalty.
}
\label{fig:identity}
\end{figure}

$G_W$ measures the penalty associated with that constraint; $L_W$ is the confusion remaining after relaxation. This is an oracle scoring diagnostic, related to local/global evaluations~\cite{plaquet2025dissecting}. It can expose time-varying slot assignments or slot reuse without distinguishing their internal causes.% Identity diagnostics merge overlapping annotations of the same speaker; the resulting DER differs from the original scoring by less than 0.0003 pp.

\begin{table}[t]
\centering
\caption{Confusion decomposition for Expert seed 0. 
Changes against Stock are pp with paired 95\% intervals.}
\label{tab:identity}

\small
\setlength{\tabcolsep}{3pt}

\begin{tabular}{lrrr}
\hline
Corpus & $\Delta C$ & $\Delta L_{30}$ & $\Delta G_{30}$ \\
\hline

\multicolumn{4}{l}{Original predictions} \\

AMI 
& +13.71 & +1.59 & +12.12 \\

EasyCom 
& +3.52 & +2.16 & +1.36 \\

PriMock57 
& -0.34 & -0.05 & -0.29 \\

\hline

\multicolumn{4}{l}{Reference-count-assisted decoding} \\

AMI
& +14.91 & +0.45 & +14.46 \\

EasyCom
& +3.75 & +1.73 & +2.02 \\

PriMock57
& -1.65 & -1.26 & -0.39 \\

\hline

\end{tabular}
\end{table}

On AMI, the added confusion is dominated by $\Delta G_{30}=12.12$ pp, compared with $\Delta L_{30}=1.59$ pp (Table~\ref{tab:identity}). Across three Expert seeds, 30-s remapping removes 88--91\% of the \emph{added} confusion. The large mapping penalty persists at 10 and 60 s and after excluding 30-s windows containing only one reference identity. At 30 s, EasyCom retains 61--67\% of its added confusion locally across seeds. This attribution is scale-dependent, for example, 10-s remapping removes 71--74\% added EasyCom confusion and 33--39\% at 30 s. Its clips are at most 60 s, so $G_{60}=0$ follows from the file length.

\subsection{Controlling speaker count}

To test whether identity degradation is explained solely by count errors, we select the top-$r_t$ output slots using the reference concurrent speaker count. Original and assisted predictions are rescored on identical eligible support (Table~\ref{tab:identity}). Assistance still leaves an AMI confusion increase of 14.91 pp from $G_{30}$ and EasyCom retains increases in both components. AMI's 10-s local confusion is lower than Stock across seeds, despite higher recording-wide confusion. We freeze each system's original mapping and compare times when both predict the correct count and found that conditional confusion increases by 19.10 pp on AMI and 4.04 pp on EasyCom (58.89\% and 50.66\% ). Instantaneous count errors alone therefore do not explain the identity deficits. %These label-assisted analyses isolate assignment behavior from output-count effects and are used only as controlled diagnostics.

\subsection{Speaker returns as a temporal probe}

On AMI, we compare the last and first up to 5 s of a speaker's preceding and returning reference turns. Both sides must contain at least 1 s where the reference and both systems have exactly one active speaker. With recording-wide mappings fixed, we measure the pre-to-post change in Expert-minus-Stock confusion. The long-minus-short contrast, for speaker inactivity of at least 30 s versus $[2,10)$ s, is positive for all three seeds: 0.98, 1.01 , and 9.98  pp. Only 20--23 long-gap events qualify from 636 candidates, supports differ across model pairs with various interval length. This suggests a conditional association with return timing, rather than establishing a general long-gap failure or a causal cache mechanism.

\section{Rehearsal gains and their cost}
\label{sec:rehearsal}

Generalist combines target-domain crops with AMI-train, AliMeeting-Eval, and MSDWild few.train. The final rule aims for equal rehearsal contributions and a 1:1 rehearsal-to-target ratio. Integer oversampling yields 92, 96, and 80 rehearsal crops versus 306 target crops per epoch and uses 3,432 optimizers (an 88.2\% increase). This comparison measures the complete rehearsal recipe under approximately matched nominal target exposure.

Generalist reduces DER relative to Expert on AMI, MSDWild, AliMeeting-2, and EasyCom. AMI's confusion reduction is primarily a decrease in $G_{30}$ with $-12.40$ pp, while the local change is $-0.49$ pp. On EasyCom, both components improve, but confusion remains 1.22 pp above Stock. Rehearsal therefore mitigates the identity deficit without uniformly restoring the original error composition. On PriMock57 the Generalist gives up 11.49 pp of ov-Miss and 10.98 pp of DER relative to Expert, eliminating the DER transfer benefit. The measured retention gains consequently come with a transfer cost and do not establish universal non-degradation.

\subsection{Pause annotation and partial recovery}
Without retraining, we merge the same speaker reference gaps and repeat matched-FA calibration. $\tau_{\mathrm{eval}}$ from 0 to 1 reverses Expert-Stock DER on MSDWild-2 (+4.99 to -2.4 pp) and AliMeeting-2 (+3.35 to -2.23 pp), see Figure~\ref{fig:tau-analysis}.

Exploratory training-label reconstruction using word timestamps and calibrated padding retains a 7.07 pp PriMock gain and 1 pp for MSDWild-2, with Alimeeting-2 still degrades of 1.18 pp. These results support annotation conventions as a partial contributor to degradation, with orpus-dependent repair costs.

\section{Limitations and conclusion}
\label{sec:conclusion}

We showed that small-data adaptation can substantially improve in-domain diarization and enable cross-domain transfer, yet detection gains coexist with degraded speaker identity consistency. Evaluation showed that speaker-count errors alone cannot explain this deficit, while local remapping distinguished local assignment errors from recording-wide consistency losses. Rehearsal mitigated degradation and largely preserved in-domain DER gains, but sacrificed transfer benefits and left residual identity degradation, demonstrating that DER recovery does not necessarily imply recovery of speaker-assignment behavior. Future work will extend the analysis to other streaming architectures, adaptations, and rehearsal strategies for balance retention and transfer.

%\vfill
%\pagebreak

\vfill\pagebreak

\bibliographystyle{IEEEbib}
\label{sec:refs}
\bibliography{strings,refs}

@misc{nvidia2025sortformerv2,
  author = {{NVIDIA}},
  title = {{Streaming Sortformer Diarizer 4spk v2}: Model Card},
  year = {2025},
  howpublished = {Hugging Face model repository},
  url = {https://huggingface.co/nvidia/diar_streaming_sortformer_4spk-v2},
  note = {[Online]. Available: \url{https://huggingface.co/nvidia/diar_streaming_sortformer_4spk-v2}. Accessed: Sept. 24, 2026}
}

@inproceedings{medennikov2025streaming,
  author = {I. Medennikov and others},
  title = {{Streaming Sortformer}: Speaker Cache-Based Online Speaker Diarization with Arrival-Time Ordering},
  booktitle = {Proc. Interspeech},
  year = {2025},
  pages = {5238--5242},
  doi = {10.21437/Interspeech.2025-2244},
  url = {https://www.isca-archive.org/interspeech_2025/medennikov25_interspeech.html}
}

@inproceedings{ryant2020third,
  author = {N. Ryant and others},
  title = {The Third {DIHARD} Diarization Challenge},
  booktitle = {Proc. Interspeech},
  year = {2021},
  pages = {3570--3574},
  doi = {10.21437/Interspeech.2021-1208},
  url = {https://www.isca-archive.org/interspeech_2021/ryant21_interspeech.html}
}

@inproceedings{park2025sortformer,
  author = {T. Park and others},
  title = {{Sortformer}: A Novel Approach for Permutation-Resolved Speaker Supervision in Speech-to-Text Systems},
  booktitle = {Proc. International Conference on Machine Learning},
  series = {Proceedings of Machine Learning Research},
  volume = {267},
  pages = {48153--48169},
  year = {2025},
  publisher = {PMLR},
  url = {https://proceedings.mlr.press/v267/park25h.html}
}

@article{kuchaiev2019nemo,
  author = {O. Kuchaiev and others},
  title = {{NeMo}: A Toolkit for Building {AI} Applications Using Neural Modules},
  journal = {arXiv preprint arXiv:1909.09577},
  year = {2019},
  doi = {10.48550/arXiv.1909.09577},
  url = {https://arxiv.org/abs/1909.09577}
}

@inproceedings{loshchilov2019adamw,
  author = {I. Loshchilov and F. Hutter},
  title = {Decoupled Weight Decay Regularization},
  booktitle = {Proc. International Conference on Learning Representations},
  year = {2019},
  url = {https://openreview.net/forum?id=Bkg6RiCqY7}
}

@inproceedings{coria2023continual,
  author = {J. M. Coria and H. Bredin and S. Ghannay and S. Rosset},
  title = {Continual Self-Supervised Domain Adaptation for End-to-End Speaker Diarization},
  booktitle = {Proc. 2022 IEEE Spoken Language Technology Workshop (SLT)},
  year = {2023},
  pages = {626--632},
  doi = {10.1109/SLT54892.2023.10023195}
}

@article{lanzendoerfer2025benchmarking,
  author = {L. A. Lanzend{\"o}rfer and F. Gr{\"o}tschla and C. Blaser and R. Wattenhofer},
  title = {Benchmarking Diarization Models},
  journal = {arXiv preprint arXiv:2509.26177},
  year = {2025},
  doi = {10.48550/arXiv.2509.26177},
  url = {https://arxiv.org/abs/2509.26177}
}

@article{plaquet2025dissecting,
  author = {A. Plaquet and others},
  title = {Dissecting the Segmentation Model of End-to-End Diarization with Vector Clustering},
  journal = {arXiv preprint arXiv:2506.11605},
  year = {2025},
  doi = {10.48550/arXiv.2506.11605},
  url = {https://arxiv.org/abs/2506.11605}
}

@inproceedings{zmolikova2024mmcsg,
  author = {K. {\v Z}mol{\'i}kov{\'a} and others},
  title = {The {CHiME-8 MMCSG} Challenge: Multi-modal Conversations in Smart Glasses},
  booktitle = {Proc. 8th International Workshop on Speech Processing in Everyday Environments (CHiME)},
  year = {2024},
  pages = {7--12},
  doi = {10.21437/CHiME.2024-2},
  url = {https://www.isca-archive.org/chime_2024/zmolikova24_chime.html}
}

@inproceedings{korfiatis2022primock57,
  author = {Papadopoulos Korfiatis, A. and Moramarco, F. and Sarac, R. and Savkov, A.},
  title = {{PriMock57}: A Dataset Of Primary Care Mock Consultations},
  booktitle = {Proc. 60th Annual Meeting of the Association for Computational Linguistics (Volume 2: Short Papers)},
  year = {2022},
  pages = {588--598},
  doi = {10.18653/v1/2022.acl-short.65},
  url = {https://aclanthology.org/2022.acl-short.65/}
}

@article{donley2021easycom,
  author = {J. Donley and others},
  title = {{EasyCom}: An Augmented Reality Dataset to Support Algorithms for Easy Communication in Noisy Environments},
  journal = {arXiv preprint arXiv:2107.04174},
  year = {2021},
  doi = {10.48550/arXiv.2107.04174},
  url = {https://arxiv.org/abs/2107.04174}
}

@inproceedings{liu2022msdwild,
  author = {T. Liu and others},
  title = {{MSDWild}: Multi-modal Speaker Diarization Dataset in the Wild},
  booktitle = {Proc. Interspeech},
  year = {2022},
  pages = {1476--1480},
  doi = {10.21437/Interspeech.2022-10466},
  url = {https://www.isca-archive.org/interspeech_2022/liu22t_interspeech.html}
}

@inproceedings{yu2022m2met,
  author = {F. Yu and others},
  title = {{M2MeT}: The {ICASSP} 2022 Multi-Channel Multi-Party Meeting Transcription Challenge},
  booktitle = {Proc. IEEE International Conference on Acoustics, Speech and Signal Processing (ICASSP)},
  year = {2022},
  pages = {6167--6171},
  doi = {10.1109/ICASSP43922.2022.9746465},
  url = {https://arxiv.org/abs/2110.07393}
}

@inproceedings{carletta2006ami,
  author = {J. Carletta and others},
  title = {The {AMI} Meeting Corpus: A Pre-announcement},
  booktitle = {Machine Learning for Multimodal Interaction},
  series = {Lecture Notes in Computer Science},
  volume = {3869},
  year = {2006},
  pages = {28--39},
  publisher = {Springer},
  doi = {10.1007/11677482_3},
  url = {https://link.springer.com/chapter/10.1007/11677482_3}
}

@article{park2022review,
  author = {T. J. Park and N. Kanda and D. Dimitriadis and K. J. Han and S. Watanabe and S. Narayanan},
  title = {A Review of Speaker Diarization: Recent Advances with Deep Learning},
  journal = {Computer Speech \& Language},
  volume = {72},
  year = {2022},
  note = {Art. no. 101317},
  doi = {10.1016/j.csl.2021.101317},
  url = {https://arxiv.org/abs/2101.09624}
}

@article{chaudhry2019tiny,
  author = {A. Chaudhry and others},
  title = {On Tiny Episodic Memories in Continual Learning},
  journal = {arXiv preprint arXiv:1902.10486},
  year = {2019},
  doi = {10.48550/arXiv.1902.10486},
  url = {https://arxiv.org/abs/1902.10486}
}

@inproceedings{rolnick2019replay,
  author = {D. Rolnick and A. Ahuja and J. Schwarz and T. Lillicrap and G. Wayne},
  title = {Experience Replay for Continual Learning},
  booktitle = {Advances in Neural Information Processing Systems},
  volume = {32},
  pages = {350--360},
  year = {2019},
  publisher = {Curran Associates, Inc.},
  url = {https://papers.neurips.cc/paper_files/paper/2019/hash/fa7cdfad1a5aaf8370ebeda47a1ff1c3-Abstract.html}
}

\end{document}